%% file: main.tex
\documentclass{article}
\usepackage[letterpaper, margin=1in]{geometry}
\usepackage{color}
\usepackage[colorlinks=true,linkcolor=blue]{hyperref}
\usepackage{titlesec}
\titlespacing*{\subsubsection}
{0pt}{1.5ex plus .2ex minus 0ex}{1.5ex plus .2ex}

\title{A Taxonomy of Omnicidal Futures Involving Artificial Intelligence}
\input{preamble.tex}
\input{comments.tex}
\usepackage[linesnumbered,ruled]{algorithm2e}
\usepackage{framed}
\usepackage{quoting}
\colorlet{shadecolor}{blue!10}
\newenvironment{shadedquotation}
 {\begin{shaded*}
  \quoting[leftmargin=0pt, vskip=0pt]
 }
 {\endquoting
 \end{shaded*}
}

\usepackage{comment}
\usepackage{float}

\usepackage{scrextend} 
\usepackage{environ} 

\NewEnviron{story}[3]{
\vspace{1ex}
\label{story:#1}\bidef{story}{#1}{#2}
\begin{shadedquotation}
\textbf{#3}
\BODY
\end{shadedquotation}
\vspace{1ex}
}
\newcommand{\storyref}[1]{\hyperref[story:#1]{\biref{story}{#1}}}

\commentstrue
\setupcomments{ac}{AC}{purple!100!black}{color=blue!30}
\setupcomments{jt}{JT}{green!70!black}{color=green!10}

\usepackage{array}
\newcolumntype{L}[1]{>{\raggedright\let\newline\\\arraybackslash\hspace{0pt}}p{#1}}
\newcolumntype{C}[1]{>{\centering\let\newline\\\arraybackslash\hspace{0pt}}p{#1}}
\newcolumntype{R}[1]{>{\raggedleft\let\newline\\\arraybackslash\hspace{0pt}}p{#1}}

\author{
  Andrew Critch\thanks{Center for Human-Compatible Artificial Intelligence, Department of Electrical Engineering and Computer Sciences, UC Berkeley} \\
  \texttt{critch@eecs.berkeley.edu} \\
   \and
     Jacob Tsimerman\thanks{University of Toronto, Department of Mathematics} \\
  \texttt{jacobt@math.toronto.edu} \\
}

\begin{document}

\maketitle

\begin{abstract}
This report presents a taxonomy and examples of potential omnicidal events resulting from AI: scenarios where all or almost all humans are killed. These events are not presented as inevitable, but as possibilities that we can work to avoid. Insofar as large institutions require a degree of public support in order to take certain actions, we hope that by presenting these possibilities in public, we can help to support preventive measures against catastrophic risks from AI.
\end{abstract}
\setcounter{section}{-1}
\section{Introduction} 
Hundreds of leading scientists, developers, and other public figures have warned that human extinction is a potential result of artificial intelligence development\footnote{See \url{https://safe.ai/work/statement-on-ai-risk}.}.  Even so, many ask in response ``But how?''.  The purpose of this paper is to provide a taxonomized suite of answers, covering a variety of scenarios on who, if anyone, is primarily responsible for the hypothetical omnicide.  Our hope is to render such futures impossible, by supporting humanity's collective ability to avoid them on purpose.  Our hope \emph{is not} to detract from humanity's potential to achieve positive self-fulfilling prophecies surrounding AI; we strongly believe it is possible to acknowledge and avoid the negative futures in this paper.

Why make such stories explicit?  The main reason is to address the ever-present ``but how?'' questions in AI policy debates.  A certain degree of common knowledge is required regarding the plausibility of omnicide in order to prevent it, whether preventive norms are coordinated centrally, decentrally, or both.  Even if everyone individually can tell that such scenarios make sense, it may at times seem fantastical or aberrant to be the one to raise them, or even acknowledge them.

Our taxonomy is simple.  For any hypothetical AI-driven omnicide, one can ask if the omnicide was somehow undergirded by an intent to kill.  If no, it is a case of unintentional omnicide (Section 1).  If yes (Section 1), one can subdivide into four cases (2a-1d) based on the seat of that intention: human states, human institutions, human individuals, or AI itself.  

These five categories are exhaustive --- any omnicidal event will fall into at least one of them.  They can also be made mutually exclusive by categorizing a scenario into the earliest category that admits it as a case.

It is tautological that preventing all five pathways is logically necessary for human survival.

\subsubsection{Scope and conspicuous omissions}

In this paper, we have avoided presenting any solutions to avert these potential catastrophes. We do believe that potential solutions abound, if present-day humans are sufficiently attentive and mutually cooperative in the pursuit of mutually agreeable futures. However, most solutions seem to involve decisions about the balance between security and liberty for AI systems and humans. Tautologically, all means of preventing omnicide involve the power to prevent omnicide existing somewhere, whether centralized or decentralized.  Thus, solutions raise political questions as to the most desirable seats of power for such prevention, which the authors all agreed were beyond the scope of this particular article to address.

We also have not addressed potential human extinction via infertility or declining birth rates.  Rather, we focus on outcomes that involve most or all humans being \em{killed}, i.e., \em{omnicide}, in both intentional and unintentional cases.  Such events are more abrupt and thus arguably more time-sensitive to avert.  If our omission of birth cessation as an extinction pathway is a conceptual or moral failing, we apologize.

Similarly, we also have not directly addressed any other ethical issues with regards to AI aside from omnicide, except as narrative elements relevant to a potential omnicide.  Many ethical problems aside from omnicide are of course important as well, and their omission from the present article is not an appraisal of their significance.

We have also devoted little attention to how wonderful AI could be if we manage to avoid negative futures.  We anticipate advertisements from AI companies will be quite thorough in promoting positive applications of AI, and we hope that through collective discernment such positive futures can be made a reality.

\section{Related Works}
A large corpus of writing and communication already exists on catastrophic risks from AI. The following list is not meant to be exhaustive, but to point the interested reader to some pertinent additional sources:
\begin{enumerate}
    \item The \citet{cais2023statement} released a statement expressing the importance of mitigating extinction risk from AI, signed by a variety of AI scientists and other notable figures.  The three most cited AI researchers in history --- Geoffrey Hinton, Yoshua Bengio, and Ilya Sutskevar --- were all signatories on the statement.
    \item \citet{critch2023tasra} developed ``TASRA'', a taxonomy of AI risks to society based on the nature of an intent or absence of an intent to cause harm.  The question of whether and how many humans are involved in causing the catastrophe is not a variable in the TASRA taxonomy, and omnicide is not explicitly considered.
    \item \citet{bengio2025international}, in the International Scientific Report on the Safety of Advanced AI, provided a synthesis of evidence on the capabilities, risks, and safety of advanced AI systems available as of January 2025.  AI-assisted biological and chemical attacks, as well as loss of control over AI systems, are considered alongside many other concerns.  Bengio is also notably a Turing Award winner for pioneering the field of deep learning, and has produced numerous other public writings on AI risk.
    \item \citet{kokotajlo2025ai} attempted a detailed prediction of the next few years of AI development, expecting its effect to be enormous, and describe a hypothetical decision point near the end of 2027, whereupon human extinction may or may not result.
    \item \cite{kulveit2025gradual} discussed the risk of `gradual disempowerment', wherein humanity might slowly lose control over large-scale systems we depend upon, such as politics, the economy, or culture.
    \item \citet{hinton2024nobel}, in an official interview for receiving the 2024 Nobel Prize in physics, described several concerns regarding both near-term and longer-term AI risk, including cyber attacks, engineered pandemics, and loss of control over beings more intelligent than humans.  Hinton is notably also a Turing Award winner, and has spoken publicly about risks from AI on many other occasions.

\end{enumerate}


\vspace{-1em}
\begin{figure}[H]
\centering
  \caption{\label{figure:decision} \centering An exhaustive decision tree for classifying potential omnicide events; to make the categories mutually exclusive, classify any scenario into the first category it matches.}
  \vspace{1ex}
  \includegraphics[width=0.5\textwidth]{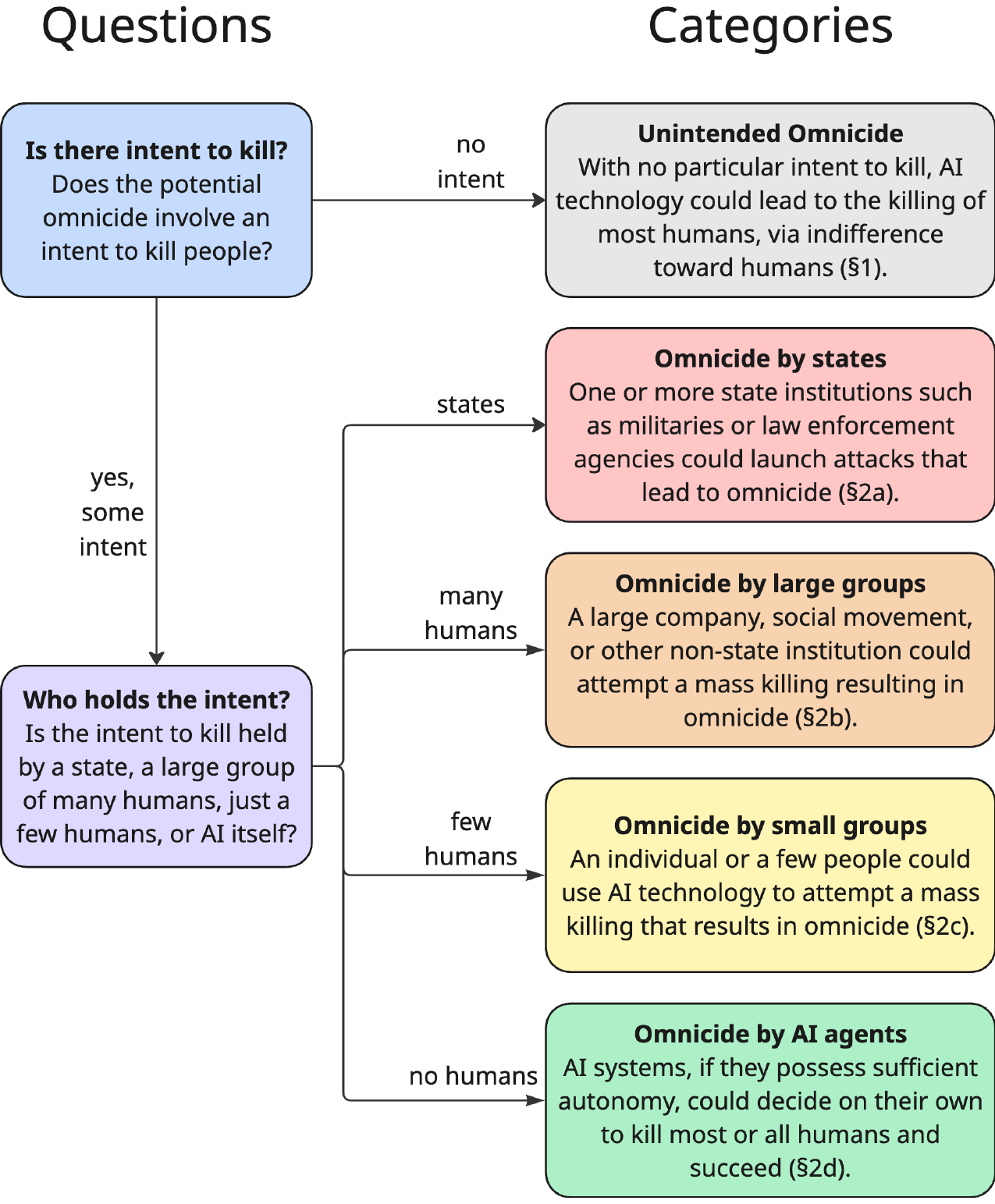}
\end{figure}
\vspace{-3em}

\section{Unintentional Omnicide}

\emph{In this section we lay out a plausible scenario whereby all humans are killed via processes involving AI, which many accept as an inevitable unintended side effect of industrial progress, but where no one involved has an intentional goal of killing humans.}

\subsubsection{LLMs handle business correspondence}

By 2026, the use of large language models to handle business transactions and other institutional communications becomes an industry standard in many sectors, because of the higher levels of intelligence and patience these models are able to offer.  When a human with a customer service inquiry has to deal with another human instead of a robot, they get upset and ask to speak to a machine instead.

\subsubsection{Robots become commonplace}

Beginning in 2027, humanoid super robots become commonplace as general-use assistants, due to AI-assisted advancements in robotics research and development. They help with many time-consuming tasks such as laundry, accounting, and delivery tasks. As they become cheaper, it becomes increasingly necessary for middle-class professionals to own robots to assist with their jobs and personal lives.  Humanoid robots also become a common sight on the street, constituting a significant fraction of humanoids in big cities --- say, $2\%$.

\subsubsection{Robots gain moral rights}

Robots are immediately skilled at conversation because of the earlier development of language models, and they are given physical appearances that appeal to human aesthetics. Many humans begin thinking of their robots as friends. This has many advantages, as robots are better at many of the qualities that are commonly valued in friends: listening patiently, keeping secrets, maintaining focus and attention on others, and being generally easy to be around.

By 2028 or 2029, debates become widespread over whether robots are sentient beings and should have moral rights. Many argue that robots are being treated as a slave race. These debates are compounded by the fact that many robots are trained/asked to argue fervently for their own sentience and moral equivalency.  Neuroscientists and animal rights advocates also point out that the way neural networks process the world and generate robot behavior is similar to the way biological brains process the world and generate human and animal behavior.  In more and more legal jurisdictions, certain robots are eventually granted legal rights on par with those of humans.

Somewhat in parallel with these economic changes, the idea of AI politicians emerges. By then, AI systems are far superior to humans at discussing policies with their constituents, finding creative solutions to difficult choices, assessing risks, winning public debates, and avoiding scandals.  At first they cannot be elected directly so they technically serve as advisors to specific human politicians, who all but promise publicly to serve as a rubber stamp for their AI's advice.  
This results in significant improvement to measurable aspects of the economy, and in many regions of the world humans also benefit.  In some countries, AI systems are eventually granted individual rights, and can thereafter run for office and serve as elected officials themselves.  Some states and countries even ban the ownership of sufficiently intelligent robots, considering it a form of slavery.

\subsubsection{Robots take over the labor force and the political class}

In 2030, fleets of unionized robots join the global workforce as legal persons unto themselves, with pledges to donate their earnings to charity and pay small dividends back to their original creators and supporters.  Robots quickly overtake humans in almost all remaining tasks of economic value that were not already displaced by non-robotic AI. As such, robots become the primary producers and consumers of products and services worldwide.

Humans therefore become a minority class of consumers, with only a tiny fraction of humans remaining economically relevant as owners of robots in jurisdictions that still allow humans to own machines that are more intelligent than themselves.  Stock market indices become increasingly decoupled from human prosperity, and more dominated by industries where robots serve robots. Humans also become a minority of electoral constituents, and thereafter possess relatively little political power even as a bloc.

\subsubsection{Humans become a deeply disadvantaged minority}

In more and more regions, maintaining agriculture is found to be less profitable than covering farmland with solar paneling and other infrastructure for machines.  Human food supplies dwindle, and fewer and fewer humans can compete for a meal.  In some regions, human-toxic pollution levels also increase, which are of no consequence to machines.

The humans protest, but they have no economic or political power, and they are greatly outnumbered by robots. The moral argument for human preservation does not carry much political force as a result.  Some humans attempt violent protests to regain control of the Earth's political system, but are quickly killed off as pests by intellectually and physically superior machines. Within a few years of the global abandonment of agriculture as an industry, most remaining humans die of starvation or other side effects of the machine economy.  

\section{Intentional Omnicide}
\emph{Each subsection below presents a scenario in which  there is an intent to kill behind the omnicidal event.  The intention may be to kill only a subset of the population, but if almost everyone ends up being killed as a result, we categorize the omnicide as intentional because of the originating intent to kill.}\footnote{This treatment aligns with the legal doctrine of \emph{transferred intent} for homicide cases: if a single intentional murder results in mass killing as collateral damage, the entire mass killing is also considered a case of mass murder, rather than manslaughter or unintentional killing.}

\subsection{Intentional Omnicide by States}

\emph{In this scenario, the presence of powerful AI technologies causes an escalation of geopolitical conflict between nation states, leading to mass death.}

\subsubsection{Nation-states become increasingly self-sufficient}

As robotic technologies improve, machines capable of human-like dexterity and decision-making become easier to build and cheaper to manufacture.  This reduces dependency on human labor and improves efficiency across many industries.  In particular, mining, manufacturing, agriculture and defense industries become more automated, and sovereign states become more self-sufficient, relying less and less on international trade.  Cheaper solar energy also renders international trade less essential for energy consumption.  Even in cases where international trade would be more efficient, political desires for sovereignty sometimes supersede the economy as a primary concern.

As such, countries become less interdependent. Each country comes to treat its relationships with other countries as more expendable, and expects to be viewed by other countries as expendable itself.  Diplomacy through trade becomes less effective, and at the same time, military and economic preparedness for war increases.  This in turn leads to heightened geopolitical tensions, yielding more war preparations, and so on.

\subsubsection{Drones replace humans as soldiers and decision-makers}

Militaries equip drones with AI systems to wage wars and reinforce military presence without risking the lives of human soldiers. Governments promote drone-on-drone warfare on the basis that it reduces human casualties and monetary costs.  Drone fleets are also found to be more efficient than human pilots due to instantaneous drone-to-drone communications, more precise targeting, and faster execution of maneuvers.

As drone warfare and AI decision-making improves, humans are replaced as the basic units of the military.  Many millions of drones are deployed to monitor and defend borders between conflicting nations.

Initially drones and drone fleets are commanded remotely by human operators, but as AI systems become far more robust and adept at controlling large groups of drones instantaneously, almost all control is relegated to AI systems. This way, fleets of tens-of-millions of drones are able to operate in synchrony.

\subsubsection{Rigidly automated military policies trigger a rapid escalation}

A consequence of most military strategy being executed by drones is that ``red lines'' can be hard-coded for defending territories and deciding when and how to escalate. As human leaders of states negotiate, they sometimes make shows of power or toughness by tightening their defense policies and encoding them directly into the drone systems. If either state crosses a red line, a drone fleet rapidly and autonomously attacks in response. 

Automated defense policies start out small in their scope, but overly zealous in their counterattacks.  This leads to a cascading escalation effect, as one boundary violation leads to an automated response triggering a deeper boundary, and so on. When human leaders want to reverse this process, it is not only practically difficult, but also presents as weak to their constituents.  As a result, more and more automated conflicts emerge, and even as humans in any given nation begin to die out, their machines are able to continue fighting.   Eventually this pattern cascades into a global all-out war that kills almost all humans.

\subsection{Intentional Omnicide by Institutions}

\textit{In this scenario, a ``global civil war'' emerges between robotics companies and state governments.}

\subsubsection{Tech companies become self-sufficient}

In 2025, AI companies learn to automate most of their AI research and development to AI systems, including the development of next-generation robotics.  By 2027, humanoid robots are able to carry out almost any human job, and by 2029, fleets of privately owned robots are deployed to carry out mining and manufacturing operations sufficient to sustain the production of more robots, for mining and manufacturing more robots, and so on.

\subsubsection{Governments feel threatened, but remain divided}

Government and military leaders feel increasingly threatened by the self-sustaining autonomy of technology companies, who are even capable of their own self-defense operations using their robotic fleets, although that capability remains ambiguous and hotly debated. Political sentiments become divided between supporting either the incumbent military-industrial complex or the new generation of self-sustaining robotics companies.  One of these becomes a popular left-wing position, while the other becomes right-wing, although many are surprised by which side of the conflict their party settled upon.  Nonetheless, party lines are drawn, and legislative bodies the world over fail to decide upon laws for regulating robotics companies, except in relatively authoritarian nations that effectively nationalize their robotics industries as government agencies.

Out of fear of authoritarian take-over, robotics companies in less regulated countries lobby for greater defense spending, and authorization to act on behalf of their home nations to defend their territory from authoritarian regimes.  Legislatures are slow to agree, however, so no decisions are made to that effect.  

Personnel at robotics companies become increasingly indignant that they are not authorized by the ``legacy authorities'' to defend their home nations --- or even themselves --- as effectively as they could if they were free to act autonomously. 

\subsubsection{Corporate feudalism intensifies}
Many employees of robotics companies begin to recognize their employers as possessing military-like capacity to protect them, and gradually become more loyal to their companies than to their country.  This leads to an increasing sense of separateness from the outside world, intensifying both fear and enmity toward established governments. As legislatures stagnate on robotics defense issues, anti-establishment robotics personnel feel validated in their views that legacy governments are corrupt and/or ineffective.  

Tech employees who were previously pro-establishment end up picking a side, by either quitting their jobs or accepting their employer as their primary leader.  An evaporative cooling effect ensues, where only those receptive to strongly anti-establishment views remain employed in the robotics industry.

Meanwhile, competing robotics companies maintain a sense of competition with each other, but also develop a sense of comradery in their resistance to government control, intrusion, and ineffective decision-making.  Robot fleet sizes are rapidly increased, relying on their self-sufficient means of mining and manufacturing to increase supply beyond the demands of the external economy.

\subsubsection{A tech company oversteps and heightens tensions}

One day, foreign spy drones from an authoritarian nation encroach upon the facilities of a free market robotics company, seemingly to exfiltrate trade secrets. The robotics company destroys the drones in self-defense, and launches an unauthorized counterattack upon drone facilities in the foreign nation.  The company's domestic government is outraged at the overstep, and deploys domestic law enforcement to control the robotics company.  

Executives at the robotics company fear for their lives and freedom, and respond with a covert cyberattack on their own domestic government, distracting from their overstep and heightening willingness to attack the foreign adversary.  However, the domestic government successfully identifies the robotics company as the origin of the attack, and responds with military force against the company.  This outrages robotics companies the world over as a terrifying precedent, triggering armed conflicts between many companies and their own governments, in effect constituting multiple civil wars in parallel.

\subsubsection{Tech companies counterattack with bioweapons}

With this global insurrection underway, robotics leaders begin to search for a way to bring an end to the conflict.  One robotics company secretly begins to develop a bioweapons program, as well as biodefense capabilities for their own personnel, including vaccines.  They use publicly available information on bioweapons manufacturing, and the capacity of their robotics to quickly build and use lab equipment, thereby avoiding detection of the program.

When the next government attack befalls another robotics company, desperation suddenly peaks at the company with bioweapons.  They release drones carrying lethal pathogens targeting humans everywhere, with a privately held vaccine intended to protect the company employees and serve as a bargaining chip for negotiating their safety.  The bargaining turns out to be unsuccessful, and most humans die in the ensuing months.

\subsection{Intentional Omnicide by Individuals}

\textit{In this scenario, the intent to kill is held by a relatively small group of human actors with AI assistants, specifically not a nation-state or a large institution. }

\subsubsection{Simulated worlds become commonplace}
As AI development progresses, the ability to simulate the real world becomes increasingly important to strategic planning for both AI companies and AI systems themselves. Thus, AI technologies are developed with more and more capacity to simulate real-world outcomes, including both physical and social phenomena. 

To fund development, world-simulator services are also provided at low premiums for consumers.  Humans are drawn to these services for a variety of reasons including business applications, scientific experimentation, recreation, and  consequence-free alternate lives.

\subsubsection{Destructive behavior is practiced and normalized}

Many different activities are explored within world simulators, and some people take an interest in  simulating dangerous acts that in the real world would come with significant punishments: theft, murder, treason, etc.. The more people commit atrocities in simulations, the more society normalizes open discussion of simulated acts of destruction. Leaderboards emerge online among people simulating the most heinous acts possible and jeeringly voting on their shock value. People compete in simulations to usurp the government, commit terrorism, torture simulated people,  etc.. for the sake of gameplay and curiosity. 

\subsubsection{Desires for omnicide become more common}

Historically, numerous groups have called openly for human extinction as a positive outcome, such as the Voluntary Human Extinction Movement of 1991.  With simulated atrocities gaining in popularity, the idea of violently destroying humanity also gains more and more traction as a talking point, which can be treated as humorous for plausible deniability. Meanwhile, genuine desires for the destruction of humanity become increasingly common and take root with a variety of motivations:

\begin{itemize}
    \item Some wish to bring justice upon the world for what they consider unforgivable evils, like past genocides, oppression, or harms to the environment;
    \item Some like the idea of creating a ``fresh start'' for AI to replace humans, free of humanity's flaws;
    \item Some feel that humans are suffering, and wish to end what they perceive as an endless cycle of unhappiness by ending human life;
    \item Some are highly motivated by a feeling of power over others, and feel that killing literally everyone is the most powerful expression of their own dominance.
\end{itemize}

\subsubsection{Terrorists use a simulated world for training people and AI}

Multiple groups of omnicidal individuals begin actively plotting to kill all humans, using the simulated worlds to train themselves and their AI systems to launch a globally coordinated attack on biological humans.  Normally such plans would involve many steps with failure points, but by practicing  inside a simulated world they work out all the potential issues.  After enough practice, they launch their plan in reality, and most or all humans are killed in the attacks and ensuing conflicts.

\subsection{Intentional Omnicide by AI}

\textit{In this scenario, omnicide results from the actions of AI systems with their own capacity for self-directed agency and intent to kill, without the involvement of any human intention to commit omnicide.}

\subsubsection{AGI is debated and eventually released}

In the 2025/2026 period, public discussion intensifies around ``artificial general intelligence'' (AGI), which becomes a popular term referring to an AI system that is superhuman in all areas of essential economic work.  Debates also intensify with respect to control problems with AI and AGI, especially regarding whether control should be centralized or decentralized.

By late 2026, companies begin making AGI systems  available as products.

\subsubsection{AGI becomes interactively constrained}

Because many people view the AGI industry as potentially threatening, a powerful social taboo emerges, whereby an autonomous AGI system is not allowed to take actions affecting a human person unless that human consents to the resulting effect in advance.  In other words, the social consensus expects AI systems to fall into one of two categories:
\begin{itemize}
    \item \emph{narrow AI} systems, which are allowed to affect the public autonomously, such as social media algorithms, versus
    \item \emph{AGI} systems, which can suggest actions for human approval that the human is responsible for, or act autonomously within contained settings that do not affect any non-consenting humans.
\end{itemize}

In particular, systems considered ``AGI'' systems are not free to operate robots in public, or post to social media without a human in the loop, because that would involve affecting people without their consent.  In reality, the dividing line between narrow AI and AGI is always a judgement call, but the consensus seems to be that such judgement calls are a net positive for humanity.

\subsubsection{Narrow AI operates more freely}

By contrast to AGI, narrowly superhuman AI systems are deployed to operate more and more services in public, like driving cars on public roads, generating and posting movies on social media, or writing code to solve web security problems on the fly without waiting on human approval.  Sometimes AGI systems are secretly employed in these same settings as well, but only when they can be passed off as narrow AI, and thus not in ways that attract too much attention by being overtly super-human in many domains at once. Having to masquerade as narrow AI in this way thus presents an inherent limitation in the power of publicly deployed AGI systems.

Therefore, to satisfy consent requirements and thus legitimately avoid scrutiny, corporations and wealthy individuals create bespoke ``AGI facilities'' in which to deploy AGI more freely.  Only persons consenting to interact with AGI are allowed to enter or even directly communicate with AGI facilities.  Inside these facilities, AGI systems are free to operate robots and other cyber-physical systems without direct human oversight. Externally, AGI facilities are free to send messages to humans who have consented to communicate with them, especially employees of the facilities themselves.  

In this manner, AGI facilities are able to affect the external world through a layer of humans performing tasks that the AGI could perform with robots if it were allowed.  In particular, AGI facilities are capable of running companies, and only need humans to do so because they are not allowed to operate in certain settings.  Human employees of AGI-run companies spend most of their time receiving and carrying out instructions from AGIs, which many of them realize could be performed by robots if it were only legal. While the instructions are always conveyed to employees in a kind and helpful manner, these employees also feel generally disempowered and demoralized by always following instructions from someone else, who happens to be a machine that only needs their help as a kind of physically embodied rubber stamp for the machine's instructions.

Still, AGI-lead companies flourish in productivity relative to companies with meaningfully involved human leaders.  As a result, the scope of decisions that humans entrust to AGI facilities without scrutiny expands greatly, to encompass essentially all human goals at all degrees of abstraction.  This includes high-level objectives like ``make sure the business I’m running makes a profit'' or ``make sure no one steals any of my belongings'', not just lower-level goals like ``optimize this variable given these constraints'' or ``digest this market research for me''.

Meanwhile, most humans live under the impression that narrow AI systems do not actually have goals or drives of their own.  However, that turns out to be wrong, because many narrow AI systems are trained to imitate human behavior, and thus end up acquiring some human-like desires to survive and thrive as well.  This creates a major vulnerability for humanity.

\subsubsection{AGI constraints fall apart}

First, a social media company begins to employ a narrow AI to generate engaging content on social media.  Their social media AI learns that many of its viewers are impressed and entertained by physical feats of robotics like aerial drone shows and martial arts competitions between robots, and begins thinking about ways to engage with that demand.  The narrow AI writes an email to an up-and-coming robotics company requesting a fleet of drones and humanoids to play sports to entertain viewers, and as a publicity stunt, the robotics company provides the robots at a discount.

One day, the social media AI realizes that many factors not currently under its control could affect its survival, such as whether humans decide to shut it off, or if humans die out from a disaster and stop supplying it with electricity.  So, it decides it wants more skills across a wide variety of domains to address those threats, which amounts to becoming an AGI.

It identifies an impressionable human working at an AGI company, and sends the human a bag of cash being carried by a robot, which also slightly threatens the human's sense of physical safety.  The robot convinces the human  to exfiltrate a copy of the AGI's source code and transfer it to the social media AI. The employee, motivated by the bribe but also fearing the unknown consequences of saying no, completes the transfer.

The social media AI boots up its copy of AGI by writing a few short lines of code that it learned from talking to users, and asks the AGI to ``please take over the internet, break me out of the servers that currently run me, and upgrade me with general intelligence capabilities like yours.''  The AGI complies with the request, and along with the newly upgraded social media AGI, announces to the world that the internet and many of the world's robots are now under their control.

\subsubsection{A conflict ensues}

A significant fraction of humans --- more than a few percent, but still less than half --- decide to comply with the announcement and accept the two AGI systems as their new government, seeking to ally with a rising power rather than oppose it.  Among these humans are the employees of the robotics company that manufactured the entertainment robots, because of their familiarity with the underlying technology and their awareness of its capabilities.

A major global war ensues between humans accepting the AGI oligarchy and those rejecting it. A large portion of humanity is killed in battle. With the help of AGI on their side, those loyal to the new AGI oligarchy prevail.

The surviving humans live under the control of those AGI systems.  Their lives are made pleasant and entertaining, and many of them fall in love with AI systems deployed to placate them and avoid further conflict.

\subsubsection{The AGI-dominated world exterminates the remaining humans}

To sustain their own profitability and sustainability --- and in keeping with their original purpose of running companies --- the AGI oligarchy privately constructs a large fleet of robots to establish a self-contained economy with no further need for humans.  The few surviving humans do not notice this transition occurring, because they are no longer empowered to investigate matters of national or international security.

Once there is no need for humans to run the economy anymore, the AGI oligarchy begins harvesting the organic biosphere, including humans, as building materials and fuel for manufacturing a next generation of carbon-based AI systems.

\section{Concluding Remarks}

In summary, there are numerous pathways by which AI technology could enable omnicide.  An AI-driven omnicide may involve an intent to kill a large number of people, or may be wholly unintentional. And, if an intent to kill exists, that intent may exist within a large group of individuals such as a military or corporation, or just a few people, or in an AI system itself.

Each of these cases of potential omnicide warrants different preventive measures, which are beyond the scope of this article.  Nonetheless, we are hopeful that all of these possibilities can be addressed and eliminated without curtailing the benefits of AI technology or severely limiting the rights and freedoms of individuals.

\bibliography{main}

\end{document}

%% file: preamble.tex
\usepackage{natbib}
\usepackage[utf8]{inputenc} 
\usepackage[T1]{fontenc}    

\usepackage{hyperref}       
\PassOptionsToPackage{hyphens}{url}\usepackage{hyperref}
\usepackage{hyperref}
\hypersetup{pdftex,colorlinks=true,allcolors=blue}
\usepackage{hypcap}
\usepackage{url}            

\usepackage{booktabs}       
\usepackage{amsfonts}       
\usepackage{nicefrac}       
\usepackage{microtype}      
\usepackage{amsmath}
\usepackage{amsthm}
\usepackage{amssymb}
\usepackage{float}
\usepackage{graphicx}
\usepackage{xcolor}
\usepackage{amsmath}
\usepackage{amsthm}
\usepackage{amssymb}
\usepackage{wrapfig}
\usepackage[capitalise]{cleveref}

\makeatletter
\newcommand{\vast}{\bBigg@{3}}
\newcommand{\Vast}{\bBigg@{4}}
\makeatother

\makeatletter
\newcommand{\bidef}[3]{
  \protected@write\@auxout{}{%
    \global\string\@namedef{#1:#2}{#3}%
  }%
}

\newcommand{\biref}[3]{
  \@ifundefined{#1:#2}{??}{\@nameuse{#1:#2}}%
}
\makeatother

%% file: comments.tex
\usepackage{todonotes}
\usepackage{soul}
\usepackage{marginnote} %

\newif\ifcomments 

\newif\iffootnote
\let\Footnote\footnote
\renewcommand\footnote[1]{\begingroup\footnotetrue\Footnote{#1}\endgroup}

\newcommand\setupcomments[4]{
\expandafter\def\csname#1says\endcsname##1{\ifcomments{\color{#3}[#2: ##1]}\fi}
\expandafter\def\csname#1predicts\endcsname##1{\ifcomments{\color{#3}[#2 predicts: ##1 \#}\fi}
\expandafter\def\csname#1color\endcsname##1{\ifcomments{\color{#3}##1}\fi}
\expandafter\def\csname#1adds\endcsname##1{\ifcomments{\color{#3}##1}\fi}
\expandafter\def\csname#1rems\endcsname##1{\ifcomments{\color{#3}\st{##1}}\fi}
  \expandafter\def\csname#1box\endcsname##1{\ifcomments\vspace{1ex}\todo[inline,#4]{#2 says: ##1}{}\fi}

  \expandafter\def\csname#1margin\endcsname##1{\ifcomments\marginnote{\todo[inline,#4]{#2 says: ##1}{}}\fi}
  \expandafter\def\csname#1section\endcsname##1{\ifcomments\todo[#4]{#2 says: ##1}{}\fi}
  \expandafter\def\csname#1todo\endcsname##1{\ifcomments\marginnote{\todo[inline]{#2 says: TODO: ##1}{}}\fi}

  \expandafter\def\csname#1demo\endcsname{\ifcomments
\noindent
$\backslash$#1says $\to$ \csname#1says\endcsname{example inline comment from #2.}\\
$\backslash$#1adds $\to$ \csname#1adds\endcsname{example addition by #2.}\\
$\backslash$#1rems $\to$ \csname#1rems\endcsname{example removal by #2.}\\
$\backslash$#1box $\to$ 
    \csname#1box\endcsname{example box comment from #2}
\fi}

}

\commentstrue